\documentclass[11pt, A4paper]{article}
\usepackage[utf8]{inputenc} 
\usepackage[T1]{fontenc}    
\usepackage{hyperref}       
\usepackage{url}     

\usepackage{fancyhdr}
\usepackage{booktabs}       
\usepackage{amsfonts}       
\usepackage{nicefrac}       
\usepackage{microtype}      
\usepackage{graphicx}
\usepackage{wrapfig}
\usepackage{color}
\usepackage{graphicx}
\usepackage{subfigure}
\usepackage{multirow}
\usepackage{diagbox}
\usepackage{amsmath,amsthm,amssymb,epsfig}

\title{\textbf{Traffic Lane Detection  using FCN}}
\date{04-04-2020 \vspace{-1em}}

\author{
  Shengchang Zhang,\qquad   Ahmed EI Koubia,\qquad   Khaled Abdul Karim Mohammed \\
  \texttt{(victorzh,\quad Akoutbia,\quad Khaledm)@stanford.edu}} 
\author{\textbf{Shengchang Zhang}$^{[1]}$,\textbf{Ahmed EI Koubia}$^{[2]}$,\textbf{Khaled Abdul Karim Mohammed}$^{[2]}$\\1- szhang63@ukt.edu,University of Tennessee, Knoxville\\2-\{Akoutbia,\quad Khaledm\}@stanford.edu, Stanford University)}

\begin{document}

\maketitle
\vspace{-2em}
\begin{abstract}
Automatic lane detection is a crucial technology that enables self-driving cars to properly position themselves in a multi-lane urban driving environments. However, detecting diverse road markings in various weather conditions is a challenging task for conventional image processing or computer vision techniques. In recent years, the application of Deep Learning and Neural Networks in this area has proven to be very effective.  In this project, we designed an Encoder-Decoder, Fully Convolutional Network for lane detection.  This model was applied to a real-world large scale dataset and achieved a level of accuracy that outperformed our baseline model.
\vspace{-1em}
\end{abstract}

\section{Introduction}
Lane detection is one of the key techniques that enables Advanced Driver Assistance System (ADAS) and autonomous driving systems  to identify lanes on the road. It provides the accurate location and shape of each lane. The lack of distinctive features can cause lane detection algorithms to be confused by other objects with similar appearance. Moreover, the inconsistent number of lanes on a road, as well as diverse lane line patterns, e.g. solid, broken, single, double, merging, and splitting lines, further hampers performance. 

We started with the initial idea of applying instance segmentation to differentiate between lanes on the road, and label them with different colors. As we moved forward, we discovered that with our existing baseline model, we were not getting good results in terms of detecting specific lane instances.  We considered applying instance segmentation, but it required more adjustments and tuning to our baseline model.  Based on the discussion we had with our instructor, we decided to apply semantic segmentation in solving our lane detection problem.
\section{Related work} \label{RelatedWork}
In recent years, many sophisticated lane detection methods have been proposed:

\textbf{Dataset processing:}
Qin Z., etc.{\cite{QinZ}}. investigated lane detection by using multiple frames from a continuous driving scene, and proposed a hybrid deep learning architecture, which combines a Convolutional Neural Network (CNN) and a Recurrent Neural Network (RNN). Han Ma, etc.,\cite{Hanma}, proposed a multi-lane detection algorithm developed based on an optimized dense disparity map estimation. However, they do not perform well in challenging and dim environments. This is particularly true when strong interference such as crossings and turnings exist. In his paper\cite{Davy}, Davy Neven etc. attempted to define the lane detection problem as an instance segmentation problem, where each lane forms its own instance that can be trained end-to-end. To parameterize the segmented lane instances, he proposed to apply a learned perspective transformation, conditioned on the image, in contrast to a fixed ”bird’s eye view” transformation.

\textbf{Models and Algorithms: }Ze Wang,\cite{Zewang} proposed LaneNet, a deep neural network based method to break down the lane detection into two stages: lane-edge proposal and lane-line localization.  However, their model detects lane lines only, and performs detections on similar lane marks on the road like arrows and characters.
Global Convolution Networks (GCN)\cite{Mahale} model used color-based segmentation to address both classification and localization issues. They offer a real-time video transfer system, which requires great deal of processing power. 
For ADAS applications, P.R.Chen\cite{Pingr} proposed a Lane Marking Detector(LMD) using a deep CNN to extract robust lane marking features by adopting dilated convolution with shallower and thinner structure to decrease the computational cost.

Several state-of-the-art approaches have achieved excellent performance in real applications. However, most methods focused on detecting the lane from one single image, and often lead to unsatisfactory performance in handling extremely challenging situations such as heavy shadow, severe mark degradation and serious vehicle occlusions.

\section{Dataset and Features} 
We had an initial plan to develop our own dataset by mounting cameras in our cars, and possibly drive in various weather and road conditions. As we explored several data collection approaches for our project, we soon realized that creating from scratch our own dataset for lane detection would a very time-consuming process.
Instead, we decided to focus the bulk of our efforts to build the Neural Networks model for this project, and use one of the online datasets.  After an extensive search, we identified two datasets for traffic lane detection: CULane\cite{CULane}and TuSimple\cite{TuSimple}. We concluded that CULane dataset was more suitable for our project because of the way it is structured and documented.  The following section provides a detailed decription of this dataset.

\subsection{Details of CULane DataSet} \label{CULane}
CULane is a large dataset for lane detection collected by cameras mounted on six different vehicles on the streets of Beijing, China.  More than 55 hours of videos were collected and 133,235 frames were extracted.  These images are not sorted and have a resolution of 1640 x 590 pixels.  The dataset is divided into 88,880 image for training, 9,675 for validation and 34,680 images for test. 
The following figure provides a breakdown of different driving conditions captured in this dataset.  Challenging driving conditions represent more than 72.3\%. Each frame is manually annotated with cubic splices. In case the traffic lanes are occluded by vehicles, the lanes are annotated based on the context. Lanes on the other side of the road are not annotated. Each image has a corresponding annotation text file providing to the x and y coordinates for key lane marking points.  
 
\begin{figure}[!htbp]
\centering    
 
\subfigure[Dataset examples for different scenarios] 
{
	\begin{minipage}{0.5\textwidth}
	\centering         
	\includegraphics[scale=0.23]{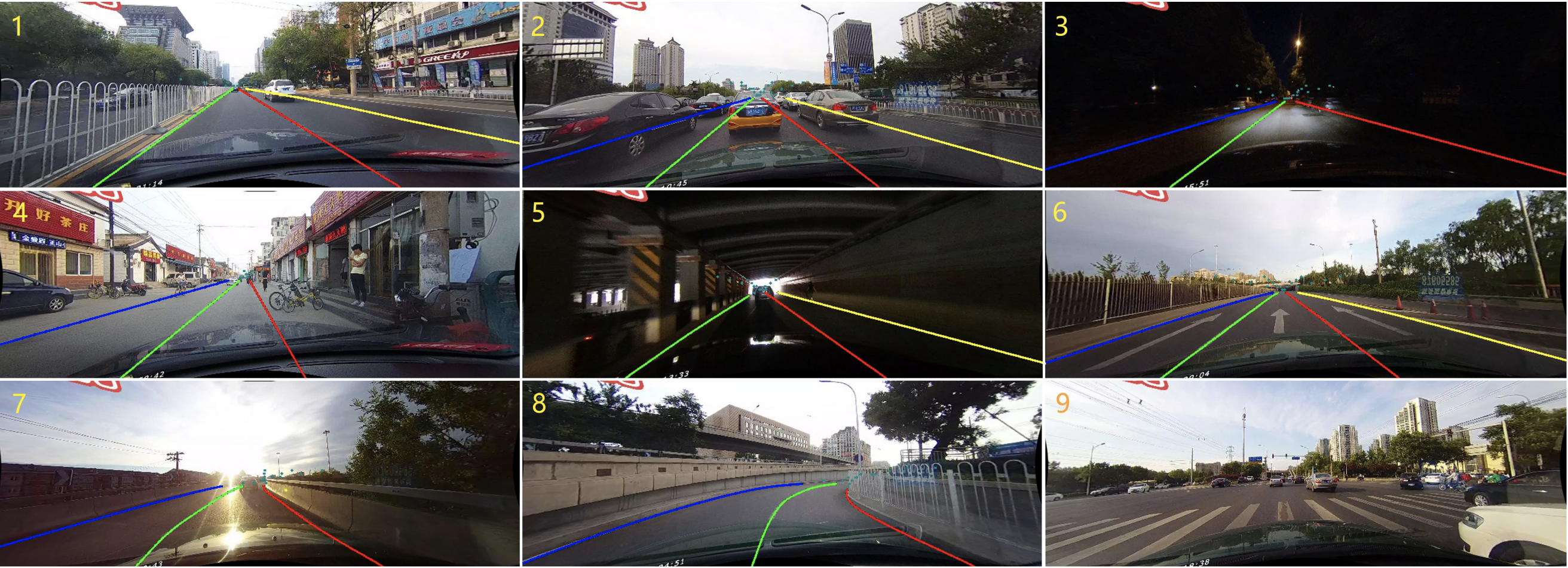}  
	\end{minipage}
}
\subfigure[ Proportion of each scenario] 
{
	\begin{minipage}{0.44\textwidth}
	\centering    
	\includegraphics[scale=0.2]{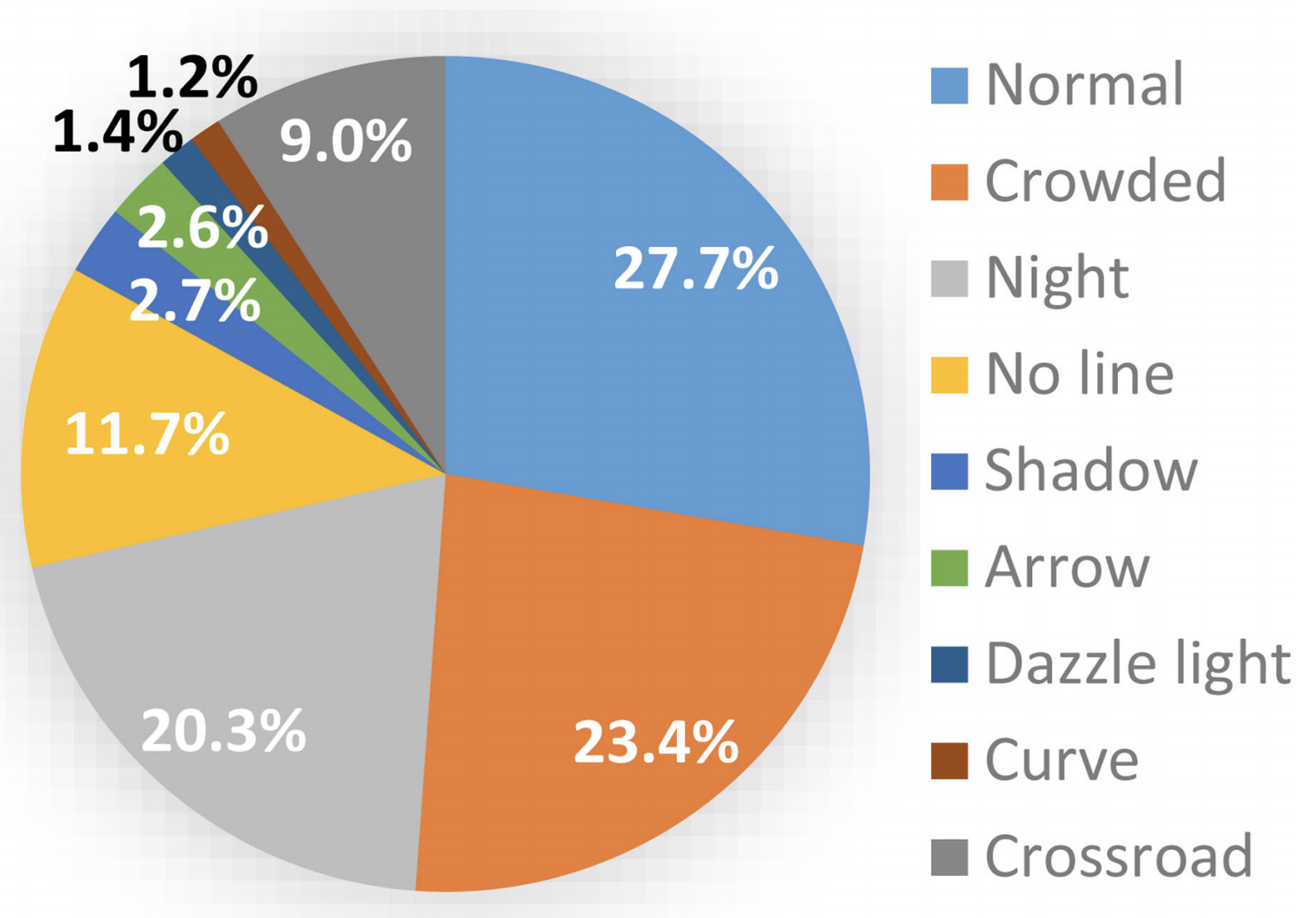}  
	\end{minipage}
}
 
\caption{CULane dataset} 
\label{fig:1}  
\end{figure}
CULane dataset is comprised of two types of images. The first consists of the original images; whereas the second category consists of mask images, which contain only the lane markings based on the original image. In case there are no lane markings, the mask image is empty. 
Each mask image contains a number representing lane markings from the original image. This number for the mask pixel denotes whether a lane is present on that particular pixel, and it takes 5 possible values. When a mask pixel has zero value of zero, it means a lane is absent. On the other hand, if a mask pixel has a value between 1 and 4, this implies that a lane is present on this pixel and the number denotes the lane number.  In short, every mask image may have from 0 to 4 lanes detected, and the lanes are numbered from left to right.

\subsection{Train \& Dev Set}
We developed a Python module that can scan the dataset folders dynamically to collect the image and label file names.  This enables us to increase or decrease the amount of data for training without any code modifications. Furthermore, we shuffled and divided the dataset it into train (90\%) and dev (10\%).

\section{Method}

\subsection{Data Pre-processing}
Initially, we developed a python script to process the dataset, and created serialized Python arrays nested in a list as expected by the baseline model, MLND-Capstone \cite{MLND}. This list contained the images as  Python arrays. The list and arrays were serialized using the Pickle API. 
We soon identified some challenges with our original pre-processing script. In particular, when we tried to load more than 600 images for training.  When all 600 training images were loaded in memory, we began to ran out of memory space.

As a result, we developed a custom Python Generator script for the dataset, which enabled us to load all training images without experiencing any memory issues. To enhance the performance further, we started loading the data in mini-batches, while configuring the optimal mini-batch size based on the GPU memory resources available. 

To speed up the training of our algorithm, we reduced the size of our images to one-fifth of their original size with maintaining the same aspect ratio. This enabled us to increase the batch size from 32 to 128, and significantly reduce the training time. As a comparison, LaneNet\cite{Davy}'s batch size is only 4 for 8 GB GPU.  Whereas, we can train our model on 8x more images for the same amount of time. 

Since we were primarily interested in identifying whether there was a lane or not in a given image, we  modified the training labels so that they can have binary values (i.e. 0 or 1 instead of the 5 values described in section \ref{CULane}).  

We also experimented with Random Channel shift as a data augmentation technique.  However, we have not seen any performance benefits resulting from applying this technique. Considering the CUlane is very large dataset and captures most the driving conditions that we were interested in, we didn't see any need for data augmentation in this project.

\subsection{Network Architecture Selection}
\begin{figure}[!htbp]
    \centering
   \includegraphics[scale=0.6] {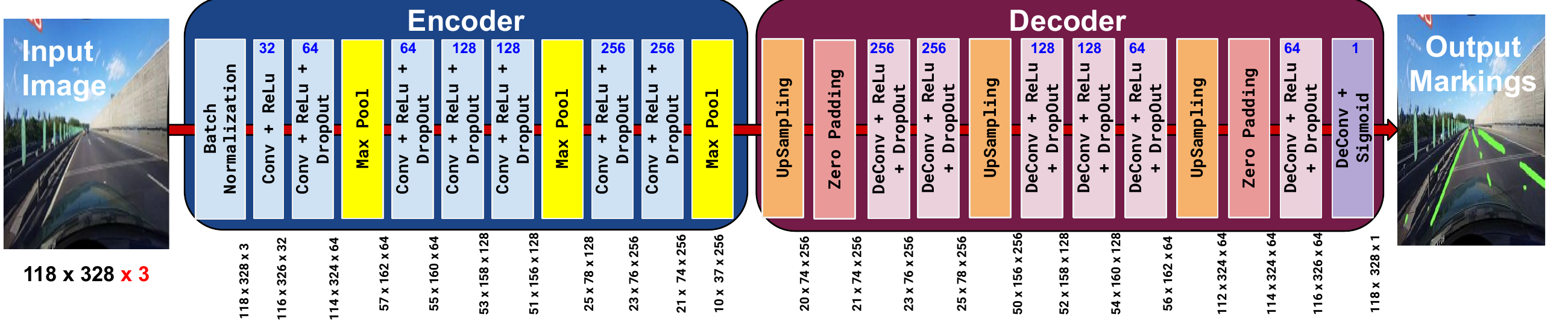}
   \caption{LaneDetect Fully Convolutional Network Architecture}\label{fig:network}
    \end{figure}

Our model has an Encoder-Decoder Network architecture, see Figure \ref{fig:network}. The Encoder portion consists of 7 Convolutional layers mixed with Dropouts and Maxpooling layers. The purpose of Encoder is to reduce the size of the the image while capturing its key features into more channels.  One the other hand, the Decorder component of the model consists of 6 layers of Deconv or ConvTranspose. It up-samples the feature vectors to an image mask, which has same height and width as the original image, on a gray-scale format.  The main idea behind choosing this network architecture is to speed up training when it comes to processing CUNet higher resolution images.  Applying alternative architectures that rely on Convolutinal Layers followed by Fully Connected Layers, would require training very large number of parameters.  In turn, this will slow down significantly the performance of our neural network.

\subsection{Baseline Model Selection}
We decided to select MLND-Capstone by Michael Virgo\cite{MLND} as the baseline model for our project because it provides good documentation and the code is easy to understand. According to Michael's project report, he was able to achieve a good performance on his dataset after 10 epoch only.

\subsection{Model Adjustments}
CULane images have a different aspect ratio compared to the baseline model's expected input resolution. To accommodate our image dimensions, we had to add 2 ZeroPadding layers to force the model to generate dimension of the label images. 
The baseline model used very few filters to capture the features of the images. This may have been sufficient for the simple two-way lane dataset, captured and labeled for the baseline model.
To achieve a good performance for CULane dataset, we had to increase the number of filters by 4x for each layer. 
    
\subsection{Loss Functions} 
The baseline model applied Keras' built-in Mean Squared Error(MSE) loss function. Since our labels are binary (i.e. 0 when a pixel is not part of a lane; and 1 when a pixel is part of a lane), we felt that it would be appropriate to use the Binary Cross Entropy (BCE) loss function. 

However, we didn't observe any performance improvements resulting from the application of the BCE loss function. After careful consideration, we realized that pixels, which are part of lanes markings represent a very small percentage compare to the total number of pixels in an image. Therefore, the labels with the value of 1 are far less than the labels with the value of 0.  In calculating the accuracy, MSE and BCE loss functions gave much bigger importance to the dominant label (i.e. label '0'). 

After some research, we discovered the Dice loss function, which provides more importance to the accuracy of label '1' (i.e the location of label '1' in predicted mask should match with the location of label '1' in the ground truth). This also ensured the label '0' to be in the right location. The Dice Coefficient is used to measure the similarity between the two samples. 

Dice Loss Function
\begin{equation} 
\label{Diceloss}
DiceLoss(p,\hat{p})=\text{1 - }\frac{2\left\langle p,\hat{p} \right\rangle }{\left| \left| p \right| \right|_{2}^{2}+\left| \left| {\hat{p}} \right| \right|_{2}^{2}}\ 
\end{equation}

\section{Experiments, Results, and Discussion}
\subsection{Hyper-parameter Tuning}
We selected the Adam Optimizer in training our model. Initially, we started with default learning rate of 0.01, and gradually lowering it until it reached a value of 0.0001. We noticed that Adam Optimizer doesn't reduce Dice loss the the learning rates are large.

\begin{figure}[!htbp]
  \begin{minipage}[t]{0.48\linewidth} 
    \includegraphics[scale=0.45]{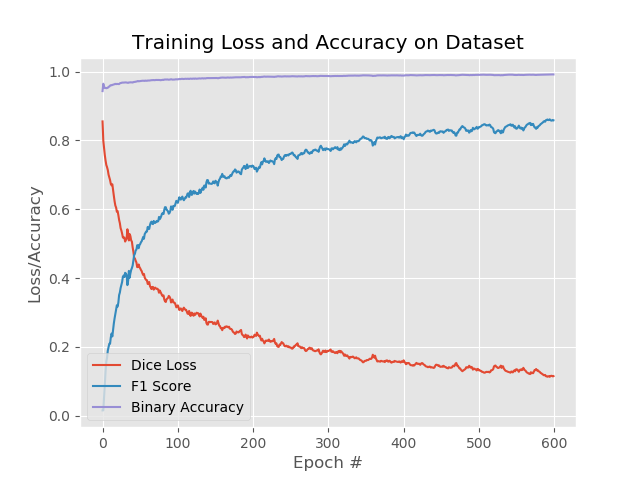} 
    \caption{Training loss and Accuracy}
   \label{fig:losstest}
  \end{minipage}%
  \begin{minipage}[t]{0.48\linewidth} 
    \includegraphics[scale=0.45]{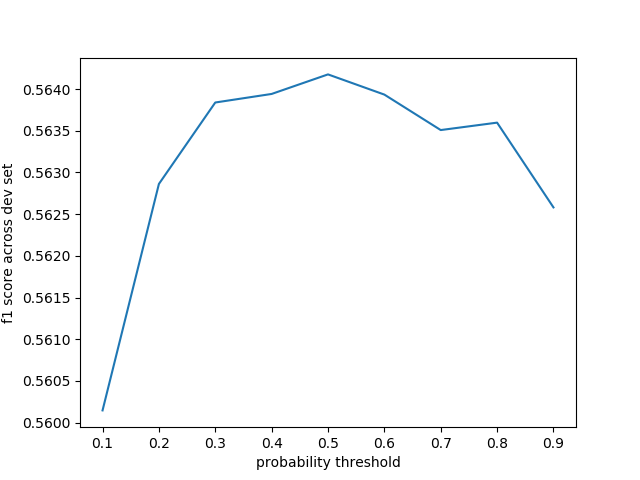} 
    \caption{F1 score vs probability threshold}
\label{fig:ROC}
  \end{minipage} 
\end{figure}

Moreover, we selected a mini-batch size of 128 image and labels. Our choice of the mini-batch size was constrained by our GPU memory. 
Since the learning rate is small, we trained our model for 600 epochs to provide sufficient time to the optimizer to reduce the loss. 
The plotted Figure \ref{fig:losstest} presents the performance metrics from our training dataset.


\subsection{Evaluation Metrics}
To evaluate the performance of our training model, we applied two metrics: F1 Score and Binary Accuracy. 
The Binary Accuracy provides 99\% accuracy, however we are not convinced that it is computing the accuracy correctly in our mask.

Even with a low learning rate, our algorithm reduced the loss to less than 20\% and we obtained an F1 score of more than 80\% within 600 epochs.  Whereas, LaneNet \cite{Davy}required 80010 epochs to achieve a goodperformance. F1 score gives us a better insight into our model’s performance. F1 is just another name of DiceLoss’ coefficient\cite{wiki}. So we can see from Figure 3 that as loss decreases, the f1 score is increasing.

\subsection{results}
\begin{figure}[!htbp]
\subfigure
{\begin{minipage}{0.02\textwidth}
(a)
  \end{minipage}
	\begin{minipage}{0.98\textwidth}
	\centering         
	\includegraphics[scale=0.395]{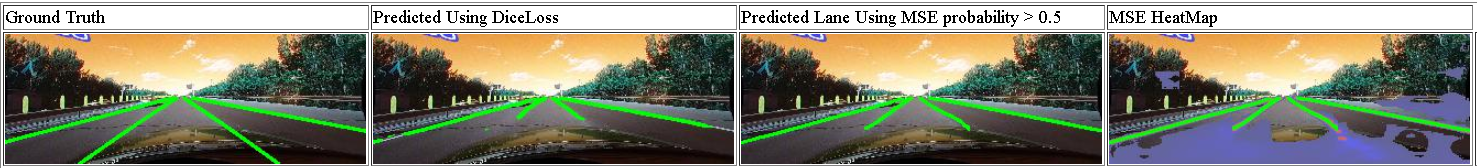}  
	\label{fig:comparision1}
	\end{minipage}
}
\vspace{-0.1in}
\hspace{-0.1in}
\subfigure
{
\begin{minipage}{0.02\textwidth}
(b)
  \end{minipage}
	\begin{minipage}{0.98\textwidth}
	\centering         
	\includegraphics[scale=0.31]{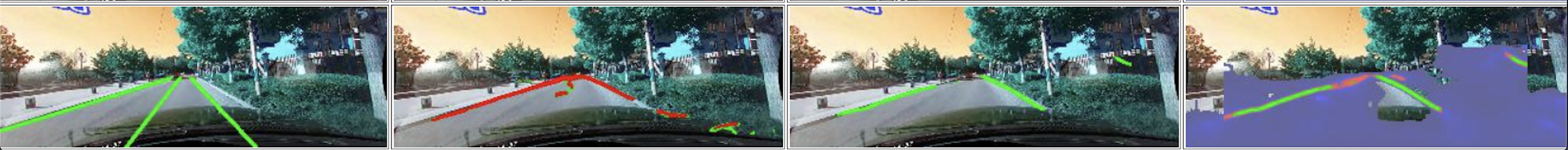}  
	
	\label{fig:comparision2}
	\end{minipage}
}
\setlength{\abovecaptionskip}{0pt} 
\caption{Lane markings results}
	\label{fig:comparision}
\end{figure}

The image  given in Fig.\ref{fig:comparision} below provides a visualized comparison of the results of the same model trained with two different loss functions. From the results, we can see that
the first row of images \ref{fig:comparision}(a) demonstrate that both loss functions achieve similar performance when detecting four traffic lanes. On the other hand,  In contrast, in some conditions the results are still not
very satisfactory, see Fig. \ref{fig:comparision}(b) the second row  demonstrates that the models are  not performing very well when it comes to a  three lanes detection with cross lanes,because lane markings usually have less appearance
clues so that dense cannot distinguish lane markings
and background.

\section{Conclusion/Future Work }

In this project we attempted to tackle the complex problem of lane detection using a large scale and challenging dataset.  We extended the baseline model which was designed initially to handle small datasets.  We developed customer Python modules to load the dataset efficiently. We tuned the hyper-parameters, extended the convolutional layers, added pooling layers and replaced the MSE cost function with the dice cost function.  This enable us to achieve a high accuracy in terms of a single lane detection.  

Further development of this project would include multi-lane detection.  In other words, for each image in the CUDataset dataset, we need to identify the lanes, their location and their unique instance (If possible, we need to detect 4 different lanes). 

To achieve this goals, we believe that it is worthwhile to pursue the following ideas:
- Instead training our neural network from scratch we will use a  use standard FCN as a backbone architecture (Alexnet, VGG, and Resnet), this will speed up the training or our model 
- Explore various Region-Based Convolutional Neural Networks (R-CNN), which 
are designed for object detection and instance segmentation. This includes Faster-RCNN and Mask-RCNN

To evaluate result of our model, we need to apply a multi-task loss function to our multi-lane detection model. This include the loss of classification, localization and segmentation mask.

\end{document}